\documentclass[preprint,12pt]{elsarticle}

\usepackage{amsmath,amssymb,amsfonts}
\usepackage{graphicx}
\usepackage{booktabs}
\usepackage{textcomp}
\usepackage{array}
\usepackage{url}
\usepackage[hidelinks]{hyperref}

\journal{Acta Astronautica}
\biboptions{numbers,sort&compress}
\graphicspath{{figures/}}

\newcommand{\R}{\mathbb{R}}

\newcommand{\Concat}{\operatorname{Concat}}
\newcommand{\Flatten}{\operatorname{Flatten}}

\newcommand{\Softmax}{\operatorname{Softmax}}

\begin{document}

\begin{frontmatter}

\title{GAP-GDRNet: Geometry-aware monocular 6D pose estimation for spacecraft using synthetic geometric supervision}

\author[hit]{Zongwu Xie}
\author[hit]{Yonglong Zhang}
\author[hit]{Yifan Yang}
\author[hit]{Yang Liu\corref{cor1}}
\ead{liuyanghit@hit.edu.cn}
\author[hit]{Guanghu Xie}

\cortext[cor1]{Corresponding author}

\affiliation[hit]{organization={State Key Laboratory of Robotics and Systems, Harbin Institute of Technology},
            city={Harbin},
            postcode={150001},
            state={Heilongjiang},
            country={China}}

\begin{abstract}
Monocular spacecraft 6D pose estimation remains difficult under weak texture, thin structures, illumination variation, and occlusion. This article presents GAP-GDRNet, a geometry-aware RGB framework built on GDR-Net for a single-target synthetic spacecraft benchmark. The method strengthens the geometry-guided regression pipeline at two points. First, AFR is placed before dense geometric prediction to combine global structural attention with local weak-texture enhancement. Second, PGSA is inserted into Patch-PnP to relate downsampled geometric regions before final pose regression. Dense supervision is obtained from a Blender-based rendering and annotation process that provides masks, model-coordinate maps, camera intrinsics, and 6D pose labels. On the self-built spacecraft dataset, GAP-GDRNet achieves a rotation error of $1.96^\circ$, a translation error of 0.0165 m, and 95.16\% ADD@0.02 m, outperforming the reproduced GDR-Net baseline by 3.88 percentage points while running at 35.97 FPS. Tests on T-LESS and LM-O further show consistent gains over the reproduced baseline on textureless and occluded non-spacecraft objects.
\end{abstract}


\end{frontmatter}

\section{Introduction}
\label{sec:introduction}

Accurate 6D pose sensing of non-cooperative spacecraft is required in rendezvous, proximity operation, on-orbit servicing, and space situational awareness. A monocular camera is attractive in these tasks because it provides compact measurements with limited hardware complexity, but the image must still support recovery of the target rotation and translation. Spacecraft differ from many terrestrial pose-estimation targets: large smooth panels, thin structural elements, repeated or symmetric parts, hard shadows, and high-contrast backgrounds often appear together. As a result, the visual evidence available from a single RGB image can be weak, unevenly distributed, or partially occluded.

\begin{figure}[t]
    \centering
    \includegraphics[width=0.92\linewidth]{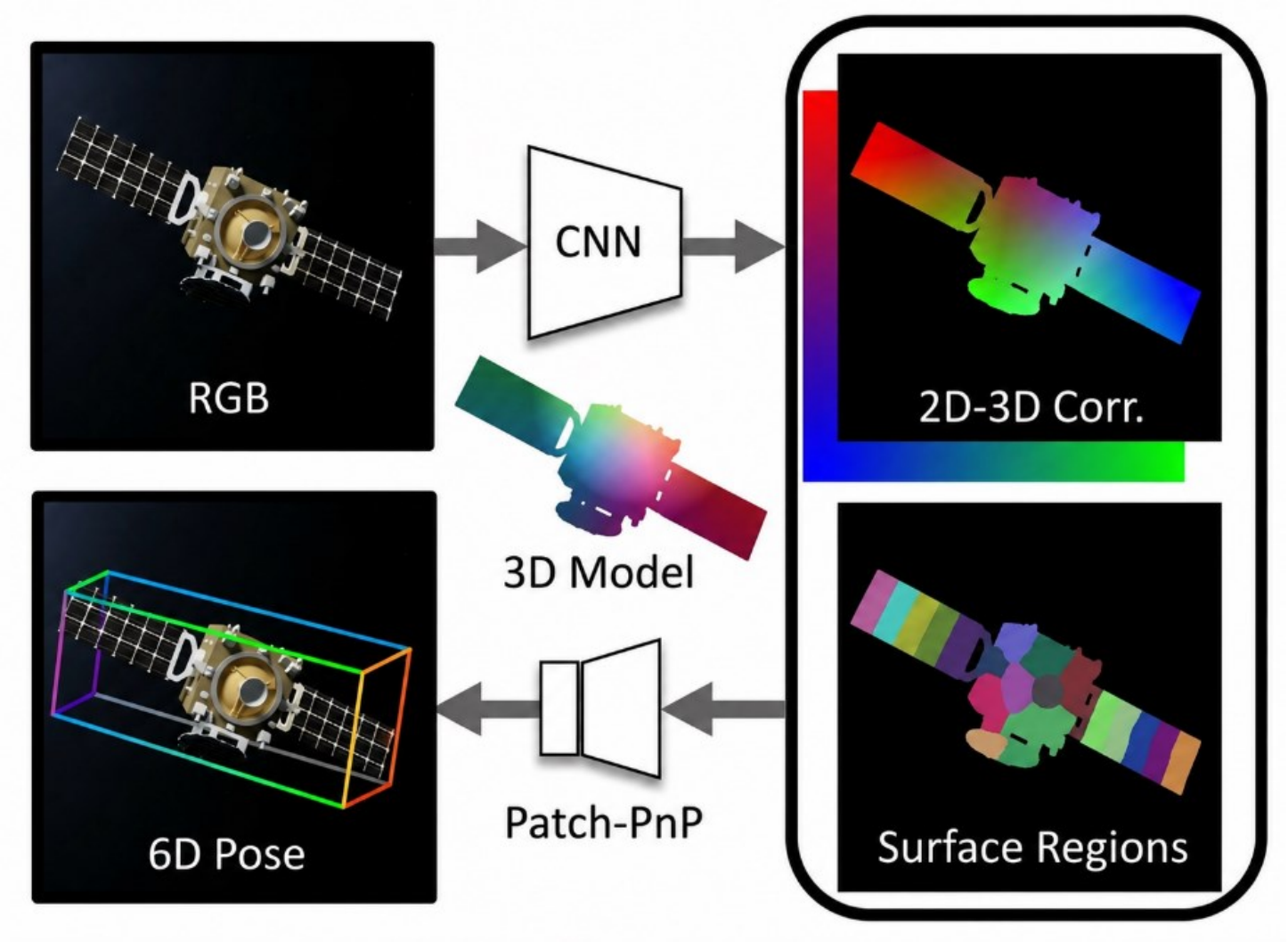}
    \caption{Geometry-guided monocular 6D pose-estimation paradigm. Dense intermediate representations, including 2D--3D correspondences and surface regions, are predicted from the RGB input and then used by Patch-PnP to regress the final spacecraft pose.}
    \label{fig:geometry_guided_paradigm}
\end{figure}

Learning-based RGB pose estimators have made dense object-specific geometry usable in monocular settings. Geometry-guided direct regression is particularly relevant here because it retains dense intermediate predictions while avoiding an external PnP solver at test time. As illustrated in Fig.~\ref{fig:geometry_guided_paradigm}, the RGB image is first converted into geometric representations such as 2D--3D correspondences and surface regions, and the final pose is then recovered by a learned Patch-PnP module. GDR-Net follows this route by predicting geometric representations from the RoI image and feeding them to Patch-PnP for final pose regression \cite{gdrnet}. For spacecraft imagery, two limitations remain. Dense coordinate and mask prediction must draw on both the coarse spacecraft layout and small boundary cues, yet illumination changes and weak texture can suppress the latter. Patch-PnP also aggregates geometric evidence after spatial encoding, where distant but related regions, such as the main body and extended panels, may not interact sufficiently through local convolutions alone.

GAP-GDRNet keeps the GDR-Net input-output setting and changes the internal representation path. Before dense prediction, attention-based feature refinement (AFR) combines a global grouped coordinate attention branch with a median-enhanced local feature branch. The former emphasizes long-range structure; the latter preserves boundary, contour, and weak-texture responses. Inside Patch-PnP, patch-level geometric self-attention (PGSA) relates downsampled geometric tokens before the final regression layers. These changes target feature formation and patch-level geometric aggregation, while inference still uses only the monocular RGB image, camera intrinsics, and target RoI.

The required dense supervision is obtained through a Blender-based rendering and annotation process, since real spacecraft images with accurate 6D pose and pixel-level geometry are scarce. The resulting dataset contains 50,000 rendered images of one target spacecraft model, with variations in pose, illumination, background, and non-target occlusion. Training, validation, and test images are split by combinations of scene factors rather than by a purely image-level random shuffle. This single-target synthetic spacecraft dataset is the main evaluation setting. T-LESS and LM-O from the BOP benchmark are used only as supplementary public datasets to check whether the same modules improve the reproduced GDR-Net baseline on textureless and occluded non-spacecraft objects \cite{bop}.

The article makes four contributions:
\begin{itemize}
    \item A monocular RGB-based spacecraft pose-sensing framework, GAP-GDRNet, is built on the GDR-Net geometry-guided regression pipeline.
    \item AFR is introduced before dense geometric prediction to strengthen global spacecraft structure and local contour cues under weak texture and structural anisotropy.
    \item PGSA is incorporated into Patch-PnP to aggregate patch-level geometric evidence before rotation and translation regression.
    \item Experiments on the single-target synthetic spacecraft dataset provide controlled comparisons with the reproduced GDR-Net baseline; T-LESS and LM-O are used as supplementary non-spacecraft tests for the same module design.
\end{itemize}

\section{Related Work}
\label{sec:related_work}

\subsection{Monocular RGB-Based 6D Object Pose Estimation}

Monocular 6D object pose estimation recovers the 3D rotation and translation of a target from a single RGB image. Learning-based methods usually follow three routes: direct pose regression, keypoint-based pose recovery, and dense correspondence prediction. PoseCNN combines object segmentation and pose prediction in cluttered scenes \cite{posecnn}. Single Shot Pose and PVNet estimate 2D keypoints or voting fields before solving for pose \cite{singleshotpose,pvnet}. Coordinate- and correspondence-based methods, including CDPN, Pix2Pose, and DPOD, predict dense geometric representations that are then converted into a 6D pose \cite{cdpn,pix2pose,dpod}. These methods show that explicit intermediate geometry can reduce the ambiguity of direct RGB-to-pose regression.

Robustness under weak texture, symmetry, and occlusion has been further studied by later RGB-based systems. EPOS handles object symmetries through fragment-based correspondences \cite{epos}; ZebraPose and SurfEmb encode surface information in different forms \cite{zebrapose,surfemb}; PFA uses perspective-flow aggregation for data-limited pose estimation \cite{pfa}. GDR-Net is closest to the present work because it predicts dense geometric representations and regresses the pose with Patch-PnP in a fully learned pipeline \cite{gdrnet}. Later geometry-guided variants further support dense intermediate geometry for RGB-based pose estimation \cite{gdrnpp}. GAP-GDRNet follows this paradigm, but changes the feature-refinement and patch-aggregation stages for weakly textured spacecraft targets.

\subsection{Spacecraft Pose Estimation and Synthetic Data}

Non-cooperative spacecraft pose estimation has been studied for rendezvous, proximity operation, active debris removal, docking, and on-orbit servicing. The visual target may contain smooth panels, thin appendages, symmetric structures, hard shadows, and large viewpoint changes; therefore, both geometric structure and appearance variation must be considered. Pauly et al. reviewed deep-learning-based monocular spacecraft pose estimation and identified dataset availability, domain gap, and deployability as central open issues \cite{acta_pauly_survey}. Neural-network-based pose estimation has also been explored for non-cooperative rendezvous \cite{spn}, while SPEED+ and SPNv2 study domain-gap-aware spacecraft pose estimation and online refinement \cite{speedplus,spnv2}.

Because real spacecraft images with accurate 6D pose and pixel-level geometry are difficult to collect, synthetic rendering and ground validation have become important tools in this field. Bechini et al. developed a validated pipeline for generating labeled spaceborne image datasets for monocular pose estimation \cite{acta_bechini_dataset}. Pasqualetto Cassinis et al. used an on-ground facility to validate a CNN-based monocular pose-estimation system under domain shift \cite{acta_cassinis_validation}. Bechini et al. further combined lightweight CNN detection with line-segment geometry for monocular relative pose initialization \cite{acta_bechini_lines}, and Rondao et al. benchmarked CNN backbones for AI-based monocular pose estimation in autonomous space refuelling \cite{acta_rondao_refuelling}. These works motivate the use of synthetic data, controlled rendering factors, and geometry-aware models. In the present work, rendering is used to supply masks, visible regions, dense model-coordinate maps, camera intrinsics, and 6D pose labels for GDR-Net-style supervision. No real captured spacecraft images are used, and the paper therefore does not claim sim-to-real spacecraft performance.

\subsection{Feature Attention and Geometry-Guided Pose Regression}

Attention modules are often used to reshape feature responses before prediction. SE and ECA recalibrate channels \cite{senet,ecanet}; CBAM, Coordinate Attention, and related spatial-channel modules combine regional emphasis with channel interaction \cite{cbam,coordatt,scse,triplet}. Non-local operations model long-range dependencies \cite{nonlocal}, and ConvNeXt provides a modern convolutional backbone \cite{convnext}. For spacecraft pose sensing, attitude ambiguity is strongly tied to global structure, while dense coordinates depend on local contours and weak-texture regions. AFR is designed around this split by refining global and local features before the geometric heads.

Final pose recovery is commonly formulated from 2D--3D correspondences with PnP solvers such as EPnP, often paired with RANSAC \cite{epnp,ransac}. GDR-Net instead uses Patch-PnP to regress pose from dense geometric predictions and RoI coordinate encoding \cite{gdrnet}. Once pose recovery is learned in this form, the aggregation of patch-level geometric evidence becomes part of the model design. PGSA addresses this step by relating patch features before regression, while the rotation output follows the continuous 6D representation \cite{rotation6d}.

\section{Method}
\label{sec:method}

GAP-GDRNet addresses monocular RGB-based 6D pose sensing for a cropped spacecraft target. Given an RGB image, camera intrinsic matrix $\mathbf{K}$, and a target bounding box, the target RoI is resized and used as the network input. The estimated pose is $\hat{\mathbf{T}}=[\hat{\mathbf{R}}\mid\hat{\mathbf{t}}]$ in the camera coordinate system, with $\hat{\mathbf{R}}\in SO(3)$ and $\hat{\mathbf{t}}\in\R^{3}$. The framework follows GDR-Net \cite{gdrnet}: dense geometric representations are predicted from the RoI image, and Patch-PnP regresses the final pose. GAP-GDRNet keeps this input-output definition and changes two internal stages, namely feature refinement before dense geometric prediction and patch-level relation modeling inside Patch-PnP. Synthetic annotations provide training supervision for dense geometry, but depth, masks, dense coordinates, and ground-truth pose are not network inputs at inference. The synthetic data process is therefore introduced first, followed by the network architecture and the two modules.

\subsection{Spacecraft Synthetic Data Generation and Annotation}
\label{subsec:data_generation}

Dense geometric supervision is needed for GDR-Net-style training, whereas real spacecraft images with accurate pose and pixel-level geometry are difficult to collect at scale. A Blender-based renderer is therefore built from a spacecraft CAD model. The CAD model is normalized to a fixed object coordinate frame and scale before rendering. Each sample stores the RGB image, camera intrinsics, target box, target-object mask, visible-region mask, dense model-coordinate map, and 6D pose label. Component-level masks are neither generated nor used.

\begin{figure}[!h]
    \centering
    \includegraphics[width=0.88\linewidth]{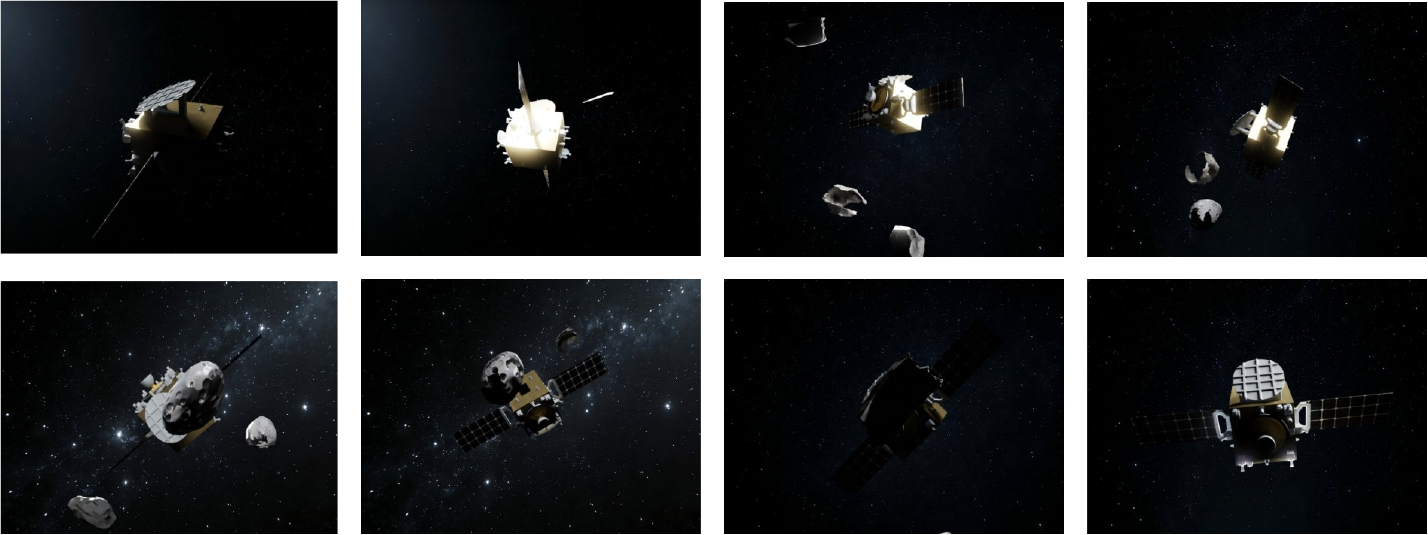}
    \caption{Examples of synthetic spacecraft images with viewpoint, illumination, background, scale, and occlusion variations.}
    \label{fig:synthetic_examples}
\end{figure}

Scene variation is introduced by sampling the camera viewpoint, target distance, relative pose, illumination direction and intensity, background image, and non-target occluding objects. Fig.~\ref{fig:synthetic_examples} shows representative samples. The rendered scenes include viewpoint and scale changes, strong and weak illumination, local shadows, star-field and Earth-like backgrounds, background clutter, and partial occlusion by other objects. The associated rendering metadata are retained for the robustness analysis in Section~\ref{sec:experiments}.

For each rendered image, the renderer records the camera intrinsics, depth buffer, and target pose $\mathbf{T}^{*}=[\mathbf{R}^{*}\mid\mathbf{t}^{*}]$ relative to the camera. The depth buffer is used only offline to create supervision. For a visible target pixel $\mathbf{p}=(u,v)$ with depth $z(\mathbf{p})$, let $\bar{\mathbf{p}}=[u,v,1]^{\top}$. Back-projection gives the camera-frame point
\begin{equation}
\mathbf{x}_{c}(\mathbf{p})
=
z(\mathbf{p})\mathbf{K}^{-1}\bar{\mathbf{p}}.
\label{eq:camera_backprojection}
\end{equation}
The corresponding object-frame coordinate is
\begin{equation}
\mathbf{C}^{*}(\mathbf{p})
=
\mathbf{R}^{*\top}
\bigl(
\mathbf{x}_{c}(\mathbf{p})-\mathbf{t}^{*}
\bigr).
\label{eq:model_coordinate}
\end{equation}
These coordinates supervise the predicted 2D--3D correspondence map $\mathbf{M}_{2D\text{-}3D}$. The visible-region mask removes self-occluded, externally occluded, back-facing, and unprojected surfaces from dense coordinate supervision.

\subsection{Overall Architecture of GAP-GDRNet}
\label{subsec:overall_architecture}

\begin{figure}[!h]
    \centering
    \includegraphics[width=\linewidth]{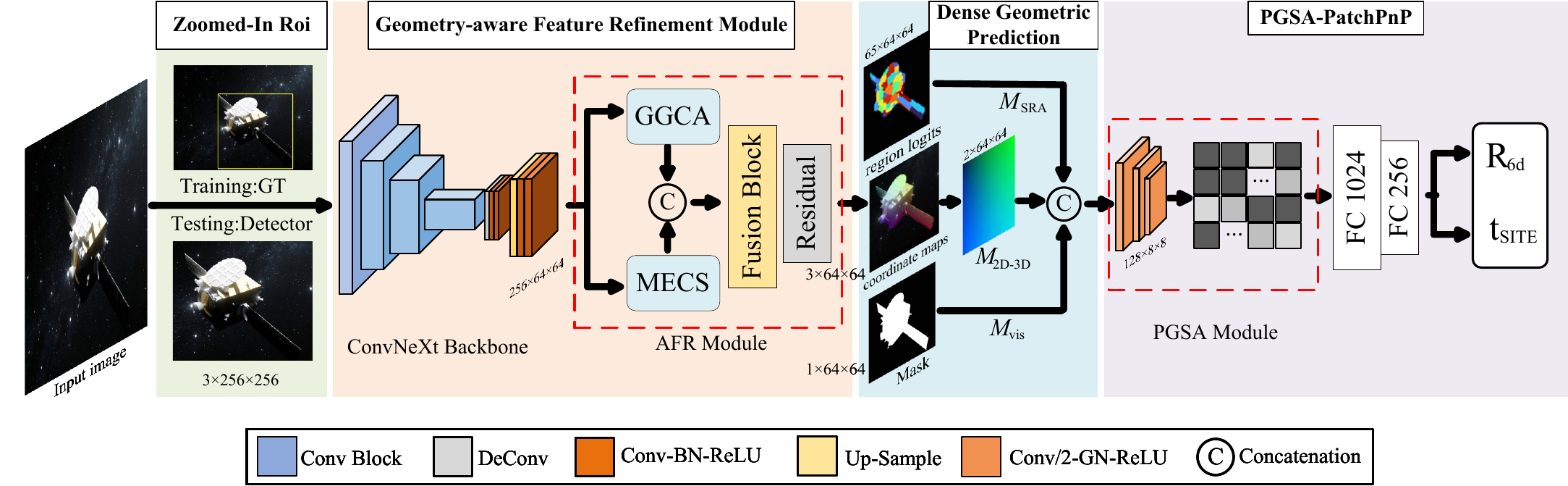}
    \caption{Overall architecture of GAP-GDRNet. The pipeline includes RoI input, ConvNeXt feature extraction, AFR feature enhancement, dense geometric prediction, and PGSA-enhanced Patch-PnP pose regression.}
    \label{fig:overall_architecture}
\end{figure}

Fig.~\ref{fig:overall_architecture} summarizes the full pipeline. GDR-Net provides the geometry-guided regression backbone \cite{gdrnet}. Relative to this baseline, AFR is placed before the dense geometric heads, and PGSA is placed in Patch-PnP before the final regression layers.

The target region is cropped from the input image and resized to a $3\times256\times256$ RoI. Depending on the experiment, the box comes from annotation or from an external detector; the RoI camera intrinsics are adjusted after cropping and resizing. A ConvNeXt backbone and decoder produce the feature map refined by AFR:
\begin{equation}
\mathbf{F}_{\mathrm{afr}}
=
f_{\mathrm{AFR}}
\bigl(
f_{\mathrm{dec}}(f_{\mathrm{backbone}}(\mathbf{I}_{\mathrm{roi}}))
\bigr).
\label{eq:overall_feature}
\end{equation}
The geometric heads predict the 2D--3D correspondence map, visible-region mask, and surface-region representation:
\begin{equation}
(\mathbf{M}_{2D\text{-}3D},\mathbf{M}_{\mathrm{vis}},\mathbf{M}_{\mathrm{SRA}})
=
\Pi_{\mathrm{geo}}(\mathbf{F}_{\mathrm{afr}}).
\label{eq:geo_outputs}
\end{equation}
The geometric outputs are concatenated with the RoI coordinate encoding to form the Patch-PnP input:
\begin{equation}
\mathbf{X}_{\mathrm{pnp}}
=
\Phi
\left(
\Concat[
\mathbf{M}_{2D\text{-}3D},
\mathbf{M}_{\mathrm{vis}},
\mathbf{M}_{\mathrm{SRA}}
],
\mathbf{U}_{\mathrm{roi}}
\right).
\label{eq:pnp_input_overall}
\end{equation}
The PGSA-enhanced Patch-PnP head then regresses the rotation representation and the scale-invariant translation estimation (SITE) representation:
\begin{equation}
(\hat{\mathbf{R}}_{6d},\hat{\mathbf{t}}_{\mathrm{SITE}})
=
\Pi_{\mathrm{pnp}}^{\mathrm{PGSA}}(\mathbf{X}_{\mathrm{pnp}}).
\label{eq:pose_outputs}
\end{equation}
The 6D rotation representation $\hat{\mathbf{R}}_{6d}$ is orthogonalized into $\hat{\mathbf{R}}$ \cite{rotation6d}. The SITE output is converted to $\hat{\mathbf{t}}$ using the centroid-depth parameterization of GDR-Net, giving the final pose $\hat{\mathbf{T}}=[\hat{\mathbf{R}}\mid\hat{\mathbf{t}}]$.

\subsection{AFR Module for Global-Local Feature Enhancement}
\label{subsec:afr}

AFR is inserted between the decoder output and the dense geometric heads. Its input is the decoded feature used for model-coordinate prediction, visible-region masking, and surface-region representation. The purpose is to strengthen the geometric evidence available before pixel-level prediction, especially the global layout of the spacecraft and the local cues around boundaries or weakly textured regions.

\begin{figure}[!h]
    \centering
    \includegraphics[width=\linewidth]{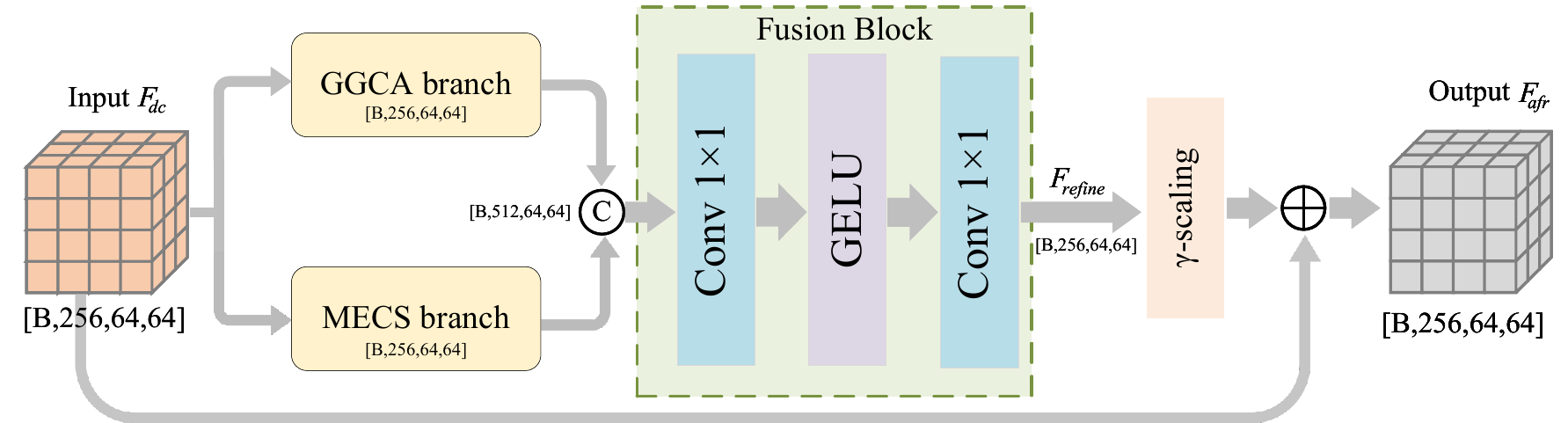}
    \caption{Overall structure of the AFR module. The decoded feature is processed by parallel GGCA and MECS branches, fused by two $1\times1$ convolution layers with GELU activation, scaled by a learnable residual factor, and added back to the input feature.}
    \label{fig:afr_overall}
\end{figure}

\noindent\textbf{A. Overall structure.}
The overall AFR structure is shown in Fig.~\ref{fig:afr_overall}. AFR contains two parallel branches: GGCA for long-range structural context and MECS for local boundary, contour, and weak-texture responses. For the decoded feature $\mathbf{F}_{\mathrm{dc}}\in\R^{B\times C\times H\times W}$, the branches produce $\mathbf{F}_{\mathrm{ggca}}=\mathcal{G}(\mathbf{F}_{\mathrm{dc}})$ and $\mathbf{F}_{\mathrm{mecs}}=\mathcal{M}(\mathbf{F}_{\mathrm{dc}})$. The two outputs are concatenated and projected:
\begin{equation}
\mathbf{F}_{\mathrm{fuse}}
=
f_{\mathrm{fuse}}
\bigl(
\Concat[\mathbf{F}_{\mathrm{ggca}},\mathbf{F}_{\mathrm{mecs}}]
\bigr).
\label{eq:afr_fuse}
\end{equation}
Residual scaling gives the refined output:
\begin{equation}
\mathbf{F}_{\mathrm{afr}}
=
\mathbf{F}_{\mathrm{dc}}
+
\gamma_{\mathrm{afr}}\mathbf{F}_{\mathrm{fuse}}.
\label{eq:afr_output}
\end{equation}
Here, $\Concat[\cdot,\cdot]$ denotes channel-wise concatenation, $\mathcal{G}(\cdot)$ and $\mathcal{M}(\cdot)$ are the GGCA and MECS transformations, and $f_{\mathrm{fuse}}$ contains two $1\times1$ convolution layers with GELU activation. The residual scale $\gamma_{\mathrm{afr}}$ is initialized to 0. Both input and output have size $B\times256\times64\times64$, so the refined feature is compatible with the original geometric heads.

\noindent\textbf{B. GGCA branch.}
GGCA targets global geometric cues in weak-texture spacecraft images. Spacecraft targets often contain elongated solar panels, a compact main body, and repeated or symmetric components. These structures may extend across a large part of the RoI, while their local texture can be weak or unevenly illuminated. A purely local convolution may therefore respond strongly to isolated edges but fail to preserve the relation between distant structural parts. GGCA is introduced to provide directional long-range modulation before dense geometric prediction.

\begin{figure}[!h]
    \centering
    \includegraphics[width=\linewidth]{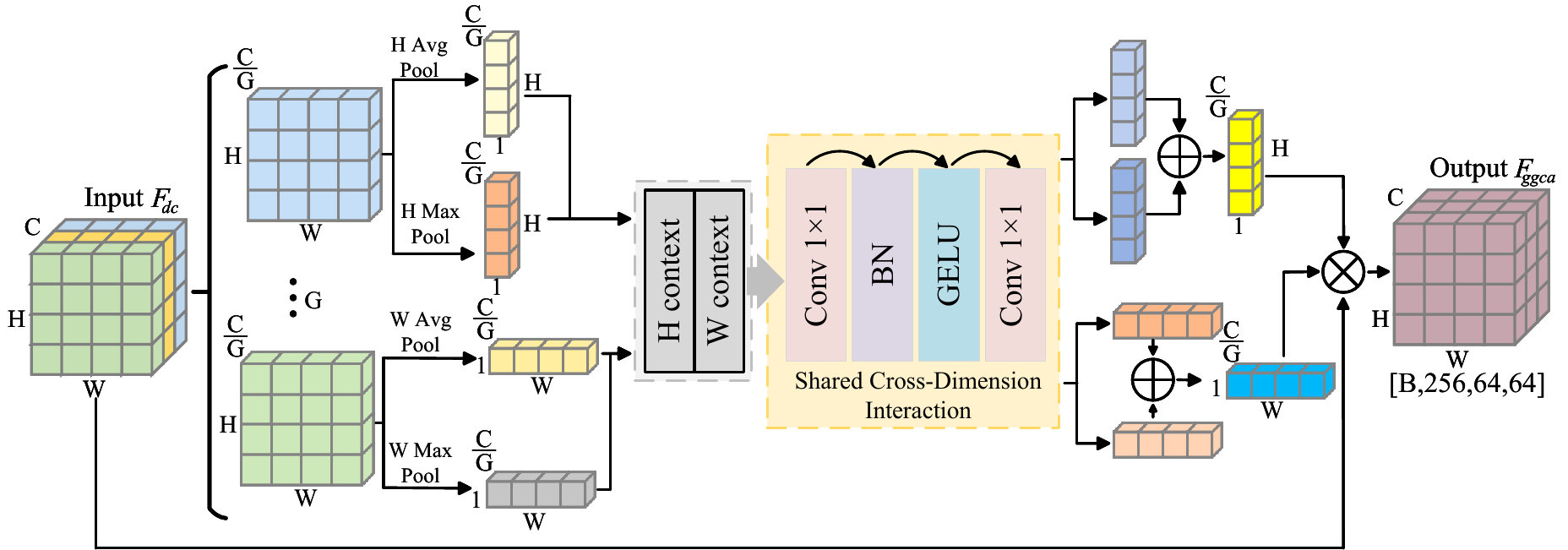}
    \caption{Structure of the GGCA global grouped coordinate attention branch. The branch uses channel grouping, horizontal and vertical directional pooling, and cross-dimensional interaction to generate directional attention for enhancing long-range spatial dependencies.}
    \label{fig:ggca}
\end{figure}

As shown in Fig.~\ref{fig:ggca}, GGCA follows the idea of coordinate attention in preserving positional information \cite{coordatt}, but uses channel grouping to avoid forcing all channels to share one attention pattern. The decoded feature is divided into $G$ groups along the channel dimension, allowing different groups to emphasize different structural directions or object parts. This design is suitable for spacecraft imagery, where the main body, solar panels, and appendages may occupy different spatial extents and orientations within the RoI.

The decoded feature is first split into $G$ channel groups:
\begin{equation}
\mathbf{F}_{\mathrm{dc}}
=
\Concat
\left[
\mathbf{F}_{\mathrm{dc}}^{1},
\mathbf{F}_{\mathrm{dc}}^{2},
\ldots,
\mathbf{F}_{\mathrm{dc}}^{G}
\right],
\quad
\mathbf{F}_{\mathrm{dc}}^{g}\in
\R^{B\times \frac{C}{G}\times H\times W}.
\label{eq:ggca_split}
\end{equation}
For the $g$-th feature group, horizontal and vertical pooling are applied separately so that the attention maps retain one-dimensional positional coordinates. Both average and max descriptors are used in each direction:
\begin{equation}
\mathbf{d}_{h}^{g}
=
\Concat
\left[
\mathrm{AvgPool}_{w}(\mathbf{F}_{\mathrm{dc}}^{g}),
\mathrm{MaxPool}_{w}(\mathbf{F}_{\mathrm{dc}}^{g})
\right],
\label{eq:ggca_h_desc}
\end{equation}
\begin{equation}
\mathbf{d}_{w}^{g}
=
\Concat
\left[
\mathrm{AvgPool}_{h}(\mathbf{F}_{\mathrm{dc}}^{g}),
\mathrm{MaxPool}_{h}(\mathbf{F}_{\mathrm{dc}}^{g})
\right].
\label{eq:ggca_w_desc}
\end{equation}
The average descriptor describes the overall structural distribution along the corresponding axis, whereas the max descriptor retains salient contour, corner, and high-response regions. The directional descriptors are passed through a shared $1\times1$ projection block $\phi(\cdot)$ and expanded to produce the horizontal and vertical attention maps:
\begin{equation}
\mathbf{A}_{h}^{g}
=
\sigma\left(f_{h}(\phi(\mathbf{d}_{h}^{g}))\right),
\quad
\mathbf{A}_{w}^{g}
=
\sigma\left(f_{w}(\phi(\mathbf{d}_{w}^{g}))\right).
\label{eq:ggca_att}
\end{equation}
The group-wise refinement is then
\begin{equation}
\tilde{\mathbf{F}}_{\mathrm{ggca}}^{g}(h,w)
=
\mathbf{F}_{\mathrm{dc}}^{g}(h,w)
\odot
\mathbf{A}_{h}^{g}(h)
\odot
\mathbf{A}_{w}^{g}(w).
\label{eq:ggca_group}
\end{equation}
\begin{equation}
\mathbf{F}_{\mathrm{ggca}}
=
\Concat
\left[
\tilde{\mathbf{F}}_{\mathrm{ggca}}^{1},
\tilde{\mathbf{F}}_{\mathrm{ggca}}^{2},
\ldots,
\tilde{\mathbf{F}}_{\mathrm{ggca}}^{G}
\right].
\label{eq:ggca_output}
\end{equation}
In this way, GGCA modulates each spatial location by two coordinate-aware factors, one along the horizontal direction and the other along the vertical direction. The branch therefore introduces directional long-range context for the target major axis, outer contour, and cross-region geometry without applying full global self-attention. The grouped design also limits parameter growth and permits part-sensitive responses in different channel subsets. The implementation uses $G=8$ channel groups and a reduction ratio of 4.

\begin{figure}[!h]
    \centering
    \includegraphics[width=\linewidth]{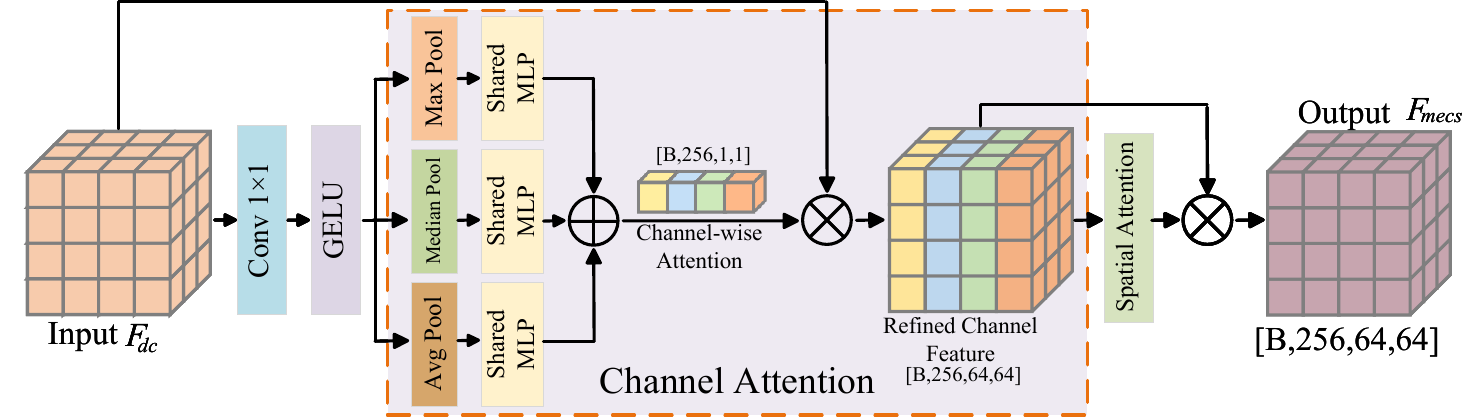}
    \caption{Structure of the MECS median-enhanced local feature branch. The branch uses max, median, and average pooling to generate channel-wise attention, and then applies spatial attention to obtain the final MECS output.}
    \label{fig:mecs}
\end{figure}

\noindent\textbf{C. MECS branch.}
MECS focuses on local geometric cues that are easily weakened by illumination changes, reflections, and low surface texture. As shown in Fig.~\ref{fig:mecs}, the decoded feature is first projected by a $1\times1$ convolution with GELU activation. Median-enhanced channel attention and spatial attention are then applied in sequence. The channel stage selects locally useful feature channels, and the spatial stage emphasizes boundaries, contours, and compact structural regions.

Given the projected feature $\mathbf{X}$, let $\mathcal{P}$ denote the three channel-pooling operators in Fig.~\ref{fig:mecs}. The channel attention map aggregates their pooled descriptors:
\begin{equation}
\mathbf{A}_{\mathrm{c}}
=
\sum_{\rho\in\mathcal{P}}
\psi(\rho(\mathbf{X})).
\label{eq:mecs_channel}
\end{equation}
\begin{equation}
\mathbf{X}_{\mathrm{c}}
=
\mathbf{A}_{\mathrm{c}}\odot \mathbf{X}.
\label{eq:mecs_channel_refine}
\end{equation}
where $\psi(\cdot)$ is a shared MLP with reduction ratio 4. The average descriptor represents the overall response level, the max descriptor keeps strong local activations, and the median descriptor reduces outlier responses caused by noise or strong reflection. After channel refinement, spatial attention highlights local boundary and contour regions. A depth-wise $5\times5$ convolution first forms the base spatial feature, and directional depth-wise convolution branches then extract multi-scale spatial responses with $k\in\{7,11,21\}$:
\begin{equation}
\mathbf{B}
=
\mathrm{Conv}_{5\times5}(\mathbf{X}_{\mathrm{c}}).
\label{eq:mecs_base}
\end{equation}
\begin{equation}
\mathbf{S}_{k}
=
\mathrm{DWConv}_{k\times1}
\bigl(
\mathrm{DWConv}_{1\times k}(\mathbf{B})
\bigr),
\quad
k\in\{7,11,21\}.
\label{eq:mecs_directional}
\end{equation}
\begin{equation}
\mathbf{S}
=
\sum_{k\in\{7,11,21\}}
\mathbf{S}_{k}.
\label{eq:mecs_spatial_sum}
\end{equation}
\begin{equation}
\mathbf{F}_{\mathrm{mecs}}
=
\sigma
\bigl(
f_{1\times1}(\mathbf{S})
\bigr)
\odot
\mathbf{X}_{\mathrm{c}}.
\label{eq:mecs_output}
\end{equation}
MECS therefore combines robust channel selection with spatial localization. The median-enhanced channel attention suppresses abnormal local responses, and the spatial attention strengthens boundary and weak-texture regions used by visible-region and dense-coordinate prediction. Together, GGCA and MECS provide complementary global and local refinement, while residual scaling keeps AFR near an identity mapping at the start of training.

\subsection{PGSA Module for Patch-Level Geometric Relation Modeling}
\label{subsec:pgsa}

\begin{figure}[!h]
    \centering
    \includegraphics[width=\linewidth]{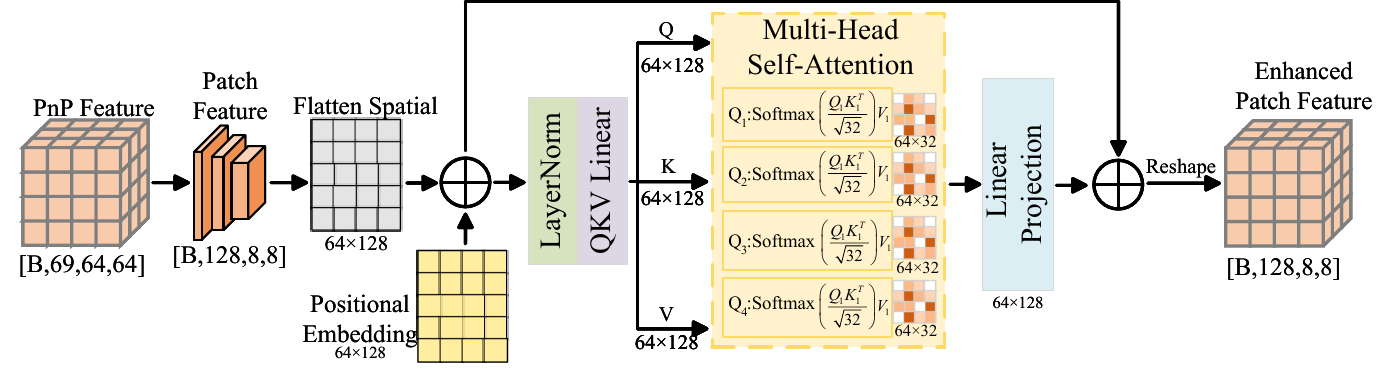}
    \caption{Structure of the PGSA module. PGSA takes the PnP feature as input, applies three stride-2 convolutions to obtain the patch feature, performs position encoding, multi-head self-attention, and residual enhancement on flattened geometric tokens, and outputs the enhanced patch feature. The subsequent two fully connected layers with 1024 and 256 units follow the baseline Patch-PnP regression head.}
    \label{fig:pgsa}
\end{figure}

PGSA is added to the Patch-PnP head after the dense geometric predictions have been assembled into $\mathbf{X}_{\mathrm{pnp}}$ and before the final fully connected regressors. It models relations among downsampled geometric regions while leaving the final rotation and translation regression form unchanged. As shown in Fig.~\ref{fig:pgsa}, the module receives $\mathbf{X}_{\mathrm{pnp}}\in\R^{B\times C_{\mathrm{pnp}}\times64\times64}$ from (\ref{eq:pnp_input_overall}).

PGSA contains local patch encoding, token-wise geometric self-attention, and residual feature restoration. Three $3\times3$ stride-2 convolution blocks with group normalization (GN) and ReLU reduce the spatial resolution from $64\times64$ to $8\times8$ and produce a 128-channel patch feature:
\begin{equation}
\mathbf{Z}
=
f_{\mathrm{s2conv}}(\mathbf{X}_{\mathrm{pnp}}).
\label{eq:pgsa_patch}
\end{equation}
The patch feature is flattened as $\mathbf{Z}_{0}=\Flatten_{\mathrm{sp}}(\mathbf{Z})$. After adding a learnable position encoding $\mathbf{P}$, LayerNorm and linear query-key-value (QKV) projection produce $\{\mathbf{Q}_{i},\mathbf{K}_{i},\mathbf{V}_{i}\}_{i=1}^{H}$. For the $i$-th head,
\begin{equation}
\mathbf{O}_{i}
=
\Softmax
\left(
\frac{\mathbf{Q}_{i}\mathbf{K}_{i}^{T}}{\sqrt{d}}
\right)
\mathbf{V}_{i},
\quad
i=1,\ldots,H .
\label{eq:pgsa_attn_map}
\end{equation}
where $H=4$ and $d=32$. The multi-head output is concatenated and projected back to the token dimension:
\begin{equation}
\mathbf{Z}_{\mathrm{att}}
=
f_{\mathrm{proj}}
\bigl(
\Concat[\mathbf{O}_{1},\ldots,\mathbf{O}_{H}]
\bigr).
\label{eq:pgsa_project}
\end{equation}
The attention feature is fused with the original patch tokens through a residual path:
\begin{equation}
\tilde{\mathbf{Z}}_{0}
=
\mathbf{Z}_{0}
+
\gamma_{\mathrm{pgsa}}\mathbf{Z}_{\mathrm{att}}.
\label{eq:pgsa_residual}
\end{equation}
The scale $\gamma_{\mathrm{pgsa}}$ is initialized to 0, so PGSA starts close to an identity mapping. The fused tokens are reshaped into an enhanced patch feature and passed to the original Patch-PnP regression head, whose two fully connected layers have 1024 and 256 units. Convolutional encoding mainly propagates local neighborhood information, but spacecraft pose often depends on separated regions such as the main-body contour, the major-axis direction, solar-panel edges, and high-curvature parts. Applying attention only to the $8\times8$ patch tokens lets PGSA model these cross-region dependencies without using self-attention on the full-resolution dense feature map.

\section{Experiments}
\label{sec:experiments}

The experiments cover the training setup, controlled comparisons, and ablations. The self-built single-target spacecraft dataset is the primary benchmark, where pose accuracy and pose-network runtime are evaluated under a common detector and test script. T-LESS and LM-O are used as public, non-spacecraft checks for textureless and occluded objects. The ablations then separate the effects of backbone replacement, AFR, PGSA, computation cost, and scene factors such as illumination, shadow, occlusion, background, and target distance.

\subsection{Experimental Setup}
\label{subsec:experimental_setup}

The primary evaluation uses the Blender-based spacecraft synthetic dataset from Section~\ref{subsec:data_generation}. It is built from one spacecraft CAD model for object-specific monocular 6D pose sensing and contains 50,000 RGB images at $640\times480$ resolution. The training, validation, and test sets follow a 70\%/10\%/20\% split. Instead of a purely image-level random split, samples are partitioned by combinations of target pose, illumination, background, and non-target occluder, so the held-out test set contains scene configurations not used for training. Each sample includes camera intrinsics, the 6D pose label, target box, target-object mask, visible-region mask, and dense model-coordinate map. Component-level masks are not used, and the depth buffer is used only during offline annotation generation.

T-LESS and LM-O \cite{bop} serve only as supplementary module-level evaluations on textureless and occluded non-spacecraft objects. They are not used to claim spacecraft-domain generalization or sim-to-real spacecraft performance.

For all experiments conducted in this article, the pose network receives a target RoI cropped from the monocular RGB image and resized to $256\times256$. The same object-level detector protocol is used for all compared pose networks. For each dataset, the detector is trained on the corresponding training split, fixed before pose evaluation, and used to generate one target box per test image. The generated boxes are stored once and reused by all reproduced methods and ablation variants, including GDR-Net and GAP-GDRNet. This keeps detection-stage variation outside the pose-network comparison. Detector runtime is excluded from the reported latency and FPS.

The implementation uses PyTorch and an NVIDIA GeForce RTX 4090 GPU. Models are trained with Ranger, i.e., RAdam with Lookahead, using an initial learning rate of $2\times10^{-4}$, batch size 24, weight decay 0.05, and 300 epochs. Training augmentation includes random brightness and contrast perturbation, background replacement, and scale perturbation. Occlusion is mainly provided by non-target objects already present in the rendered synthetic scenes.

The main module settings are fixed across all relevant experiments. AFR uses eight GGCA channel groups and a reduction ratio of 4 in both GGCA and MECS. GGCA and MECS are fused by channel concatenation followed by two $1\times1$ convolution layers with GELU activation, and $\gamma_{\mathrm{afr}}$ is initialized to 0. In PGSA, three stride-2 convolution blocks with GN and ReLU encode the Patch-PnP feature. Self-attention is applied on an $8\times8$ token grid with four heads, a 128-dimensional QKV projection, 32 channels per head, and $\gamma_{\mathrm{pgsa}}$ initialized to 0. The loss terms are
\begin{equation}
\mathcal{S}_{\mathrm{loss}}
=
\{\mathrm{coord},\mathrm{mask},\mathrm{region},\mathrm{pm},\mathrm{cent},z\}.
\label{eq:loss_set}
\end{equation}
The total objective is
\begin{equation}
\mathcal{L}
=
\sum_{s\in\mathcal{S}_{\mathrm{loss}}}
\lambda_s\mathcal{L}_s.
\label{eq:total_loss}
\end{equation}
where $\mathcal{L}_{\mathrm{coord}}$, $\mathcal{L}_{\mathrm{mask}}$, $\mathcal{L}_{\mathrm{region}}$, $\mathcal{L}_{\mathrm{pm}}$, $\mathcal{L}_{\mathrm{cent}}$, and $\mathcal{L}_{z}$ are the dense coordinate, mask, region, point-matching, center-offset, and depth losses. Their weights $\lambda_{\mathrm{coord}}$, $\lambda_{\mathrm{mask}}$, $\lambda_{\mathrm{region}}$, $\lambda_{\mathrm{pm}}$, $\lambda_{\mathrm{cent}}$, and $\lambda_z$ are set to 1.0, 1.0, 1.1, 1.0, 1.0, and 1.0.

For the self-built spacecraft dataset and ablations, the reported metrics are rotation error, translation error, and ADD@0.02 m. After CAD normalization, the 0.02 m ADD threshold is approximately 10\% of the target model diameter, so it is used as a strict tolerance for this single-target benchmark. For T-LESS and LM-O, BOP Average Recall (AR) is reported under the BOP protocol. Pose-network efficiency is measured by latency and FPS with batch size 1 after the RoI is obtained. In the main comparison and module ablation tables, bold and underline denote the best and second-best results. In the public benchmark table, highlighting is only a within-column visual aid because published rows and reproduced rows do not share identical detectors or training protocols. In the factor-wise robustness table, only the best result in each row is highlighted.

\subsection{Comparative Experiments}
\label{subsec:comparative_experiments}

The comparison is split by purpose. The self-built spacecraft dataset provides the controlled setting: the compared methods use the same detector outputs, input resolution, hardware, and evaluation script, and both accuracy and pose-network runtime are reported. T-LESS and LM-O provide public BOP references for non-spacecraft objects. On these two datasets, the controlled comparison is the reproduced GDR-Net baseline versus GAP-GDRNet; published RGB-based methods are included only as contextual references under BOP AR.

\subsubsection{Overall Performance on the Self-Built Spacecraft Dataset}

On the self-built spacecraft dataset, the comparison includes ZebraPose, MRC-Net, SurfEmb, the reproduced GDR-Net baseline, and GAP-GDRNet. ZebraPose and SurfEmb use surface encoding or embedding for RGB pose estimation \cite{zebrapose,surfemb}. MRC-Net performs direct RGB-based pose regression with multi-scale residual correlation \cite{mrcnet}. Table~\ref{tab:spacecraft_overall} reports rotation error, translation error, ADD@0.02 m, and per-image pose-network runtime under the same input resolution and hardware.

\begin{figure}[!h]
    \centering
    \includegraphics[width=0.88\linewidth]{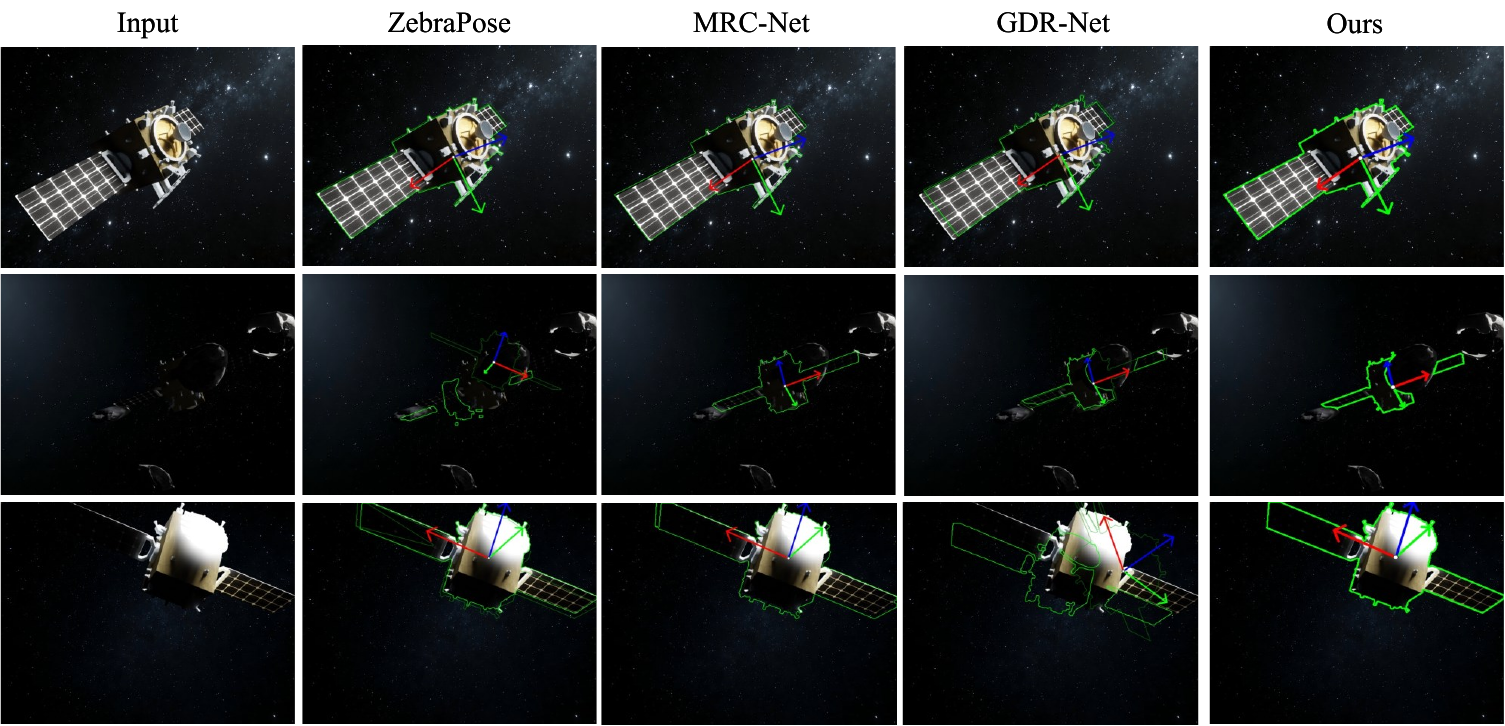}
    \caption{Qualitative pose-projection comparison on the self-built spacecraft dataset. The columns show the input image and the projected pose results of ZebraPose, MRC-Net, GDR-Net, and GAP-GDRNet under different illumination, background, viewpoint, and occlusion conditions.}
    \label{fig:qualitative_comparison_selfbuilt}
\end{figure}

Table~\ref{tab:spacecraft_overall} shows that GAP-GDRNet gives the lowest pose errors and the highest ADD@0.02 m on the self-built spacecraft dataset. Relative to GDR-Net, the rotation error decreases from $3.12^\circ$ to $1.96^\circ$, the translation error from 0.0243 m to 0.0165 m, and ADD@0.02 m increases from 91.28\% to 95.16\%. Compared with MRC-Net, the second-best method in pose accuracy, the gains are $0.47^\circ$ in rotation error, 0.0044 m in translation error, and 1.62 percentage points in ADD@0.02 m. The added modules increase latency over the reproduced GDR-Net baseline, but GAP-GDRNet still runs at 35.97 FPS and is faster than MRC-Net, ZebraPose, and SurfEmb in this setting. Fig.~\ref{fig:qualitative_comparison_selfbuilt} gives qualitative pose-projection examples under difficult appearance conditions.

\begin{table}[t!]
    \centering
    \caption{Overall performance and pose-network inference efficiency comparison on the self-built spacecraft dataset.}
    \label{tab:spacecraft_overall}
    \resizebox{\linewidth}{!}{%
    \begin{tabular}{lccccc}
        \toprule
        Method & $e_R$ (deg) $\downarrow$ & $e_T$ (m) $\downarrow$ & ADD@0.02 m (\%) $\uparrow$ & Latency / ms $\downarrow$ & FPS $\uparrow$ \\
        \midrule
        ZebraPose~\cite{zebrapose} & 2.74 & 0.0223 & 92.74 & 110.0 & 9.09 \\
        MRC-Net~\cite{mrcnet} & \underline{2.43} & \underline{0.0209} & \underline{93.54} & 68.5 & 14.60 \\
        SurfEmb~\cite{surfemb} & 2.89 & 0.0237 & 92.81 & 1121.0 & 0.89 \\
        GDR-Net~\cite{gdrnet} & 3.12 & 0.0243 & 91.28 & \textbf{22.0} & \textbf{45.45} \\
        GAP-GDRNet (Ours) & \textbf{1.96} & \textbf{0.0165} & \textbf{95.16} & \underline{27.8} & \underline{35.97} \\
        \bottomrule
    \end{tabular}
    }
\end{table}

\subsubsection{Supplementary Experiments on T-LESS and LM-O}
T-LESS mainly stresses textureless or weakly textured industrial objects, and LM-O stresses occlusion with incomplete visible target regions. These datasets test whether the same modules improve the reproduced GDR-Net baseline outside the spacecraft setting. No spacecraft-specific prior is introduced. Table~\ref{tab:public_bop} shows gains of 6.8 percentage points on T-LESS and 3.1 percentage points on LM-O over reproduced GDR-Net. These results support controlled baseline improvement on textureless and occluded non-spacecraft objects; they are not used to rank the method against all published RGB-based systems.

\begin{table}[!h]
    \centering
    \caption{Comparison with RGB-based methods on T-LESS and LM-O using the BOP Average Recall metric.}
    \label{tab:public_bop}
    \resizebox{\linewidth}{!}{%
    \begin{tabular}{lccc}
        \toprule
        Method & Single Model & T-LESS AR (\%) $\uparrow$ & LM-O AR (\%) $\uparrow$ \\
        \midrule
        CDPNv2~\cite{cdpn} & $\times$ & 40.7 & 62.4 \\
        EPOS~\cite{epos} & \checkmark & 46.7 & 54.7 \\
        DPODv2~\cite{dpodv2} & $\times$ & 63.6 & 58.4 \\
        CosyPose~\cite{cosypose} & $\times$ & 64.0 & 63.3 \\
        ZebraPose~\cite{zebrapose} & $\times$ & 72.3 & \textbf{72.1} \\
        SurfEmb~\cite{surfemb} & \checkmark & \underline{74.7} & 65.6 \\
        GDR-Net~\cite{gdrnet} & $\times$ & 70.8 & 67.2 \\
        GAP-GDRNet (Ours) & $\times$ & \textbf{77.6} & \underline{70.3} \\
        \bottomrule
    \end{tabular}
    }
\end{table}

\subsection{Ablation Studies}
\label{subsec:ablation}

\subsubsection{Module Combination Analysis}

Ablations are run on the self-built spacecraft dataset to separate backbone replacement, AFR, and PGSA. Each variant is trained and evaluated three times, and the table reports the average. The three random seeds, training schedule, detector outputs, and evaluation script are shared across variants; no row is selected from a best single run. The three-run protocol is used to reduce seed-specific variation, while the table reports means to keep the presentation compact. The baseline is GDR-Net with ResNet-34. The backbone-control variant replaces ResNet-34 with ConvNeXt while disabling GGCA, MECS, and PGSA. Other variants add GGCA, MECS, PGSA, AFR, or the full module set on the same ConvNeXt backbone.

The backbone-control row reduces the rotation error from $3.12^\circ$ to $2.91^\circ$ and raises ADD@0.02 m from 91.28\% to 92.04\%, isolating the effect of replacing ResNet-34 with ConvNeXt. Under the same ConvNeXt backbone, GGCA reduces the rotation error to $2.54^\circ$, MECS gives the lowest translation error among the single-branch variants, and PGSA raises ADD@0.02 m to 93.08\%. Combining GGCA and MECS as AFR gives $2.21^\circ$, 0.0187 m, and 94.37\% ADD@0.02 m, which indicates a positive joint effect between the two feature branches. The full GAP-GDRNet obtains the best row in the table: $1.96^\circ$, 0.0165 m, and 95.16\%. Adding PGSA on top of AFR increases ADD@0.02 m by 0.79 percentage points, suggesting that patch-level geometric relation modeling remains useful after feature refinement.

\begin{figure}[!h]
    \centering
    \includegraphics[width=0.88\linewidth]{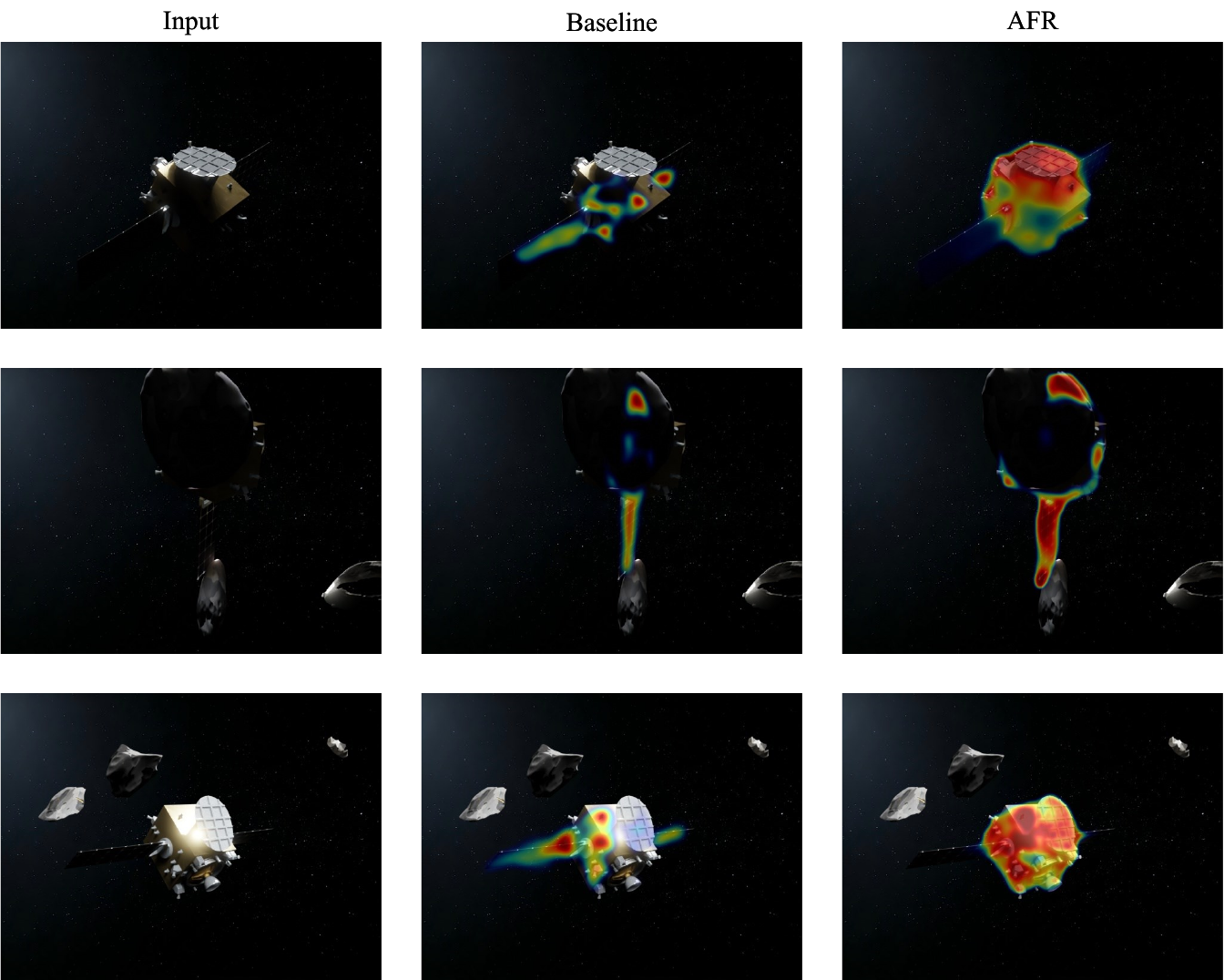}
    \caption{Feature-response visualization for the AFR variant on representative spacecraft images. The heat maps are obtained from the channel-wise $\ell_2$ norm of the feature map before dense geometric prediction and are min--max normalized for each image. Baseline denotes the corresponding model without AFR, and AFR denotes the same setting with the AFR module enabled.}
    \label{fig:afr_feature_response}
\end{figure}

Fig.~\ref{fig:afr_feature_response} inspects the feature response before dense geometric prediction. Without AFR, responses are more scattered and sometimes activate background or occluding regions. With AFR, stronger responses appear on the target body and visible structural boundaries. This visual trend is consistent with the GGCA, MECS, and AFR gains in Table~\ref{tab:ablation}. PGSA is not visualized through this feature map because it acts later, inside Patch-PnP; its effect is reflected by the PGSA and full-model rows.

\begin{table}[!h]
    \centering
    \caption{Module combination ablation results on the self-built spacecraft synthetic dataset. Each result is the average of three runs.}
    \label{tab:ablation}
    \resizebox{\linewidth}{!}{%
    \begin{tabular}{lccccccc}
        \toprule
        Variant & Backbone & GGCA & MECS & PGSA & $e_R\downarrow$ / deg & $e_T\downarrow$ / m & ADD@0.02 m (\%) $\uparrow$ \\
        \midrule
        GDR-Net & ResNet-34 & -- & -- & -- & 3.12 & 0.0243 & 91.28 \\
        Backbone-control variant & ConvNeXt & -- & -- & -- & 2.91 & 0.0230 & 92.04 \\
        + GGCA & ConvNeXt & \checkmark & -- & -- & 2.54 & 0.0225 & 92.86 \\
        + MECS & ConvNeXt & -- & \checkmark & -- & 2.73 & 0.0202 & 93.21 \\
        + PGSA & ConvNeXt & -- & -- & \checkmark & 2.57 & 0.0229 & 93.08 \\
        + AFR (GGCA + MECS) & ConvNeXt & \checkmark & \checkmark & -- & \underline{2.21} & \underline{0.0187} & \underline{94.37} \\
        GAP-GDRNet (full) & ConvNeXt & \checkmark & \checkmark & \checkmark & \textbf{1.96} & \textbf{0.0165} & \textbf{95.16} \\
        \bottomrule
    \end{tabular}
    }
\end{table}

\subsubsection{Complexity Analysis}

Table~\ref{tab:complexity} reports the computational cost of representative variants. The ConvNeXt control model is larger than GDR-Net but still reaches 41.15 FPS. Relative to this control, AFR adds 0.45M parameters and 1.06G FLOPs, and PGSA adds 0.07M parameters and 0.07G FLOPs. The full model has 47.75M parameters and 14.99G FLOPs; its latency rises from 24.3 ms to 27.8 ms compared with the ConvNeXt control. The final pose-network speed is 35.97 FPS in this setting.

\begin{table}[!h]
    \centering
    \caption{Complexity analysis of representative variants on the self-built spacecraft dataset. Pose-network latency and FPS are measured with batch size 1 on the same GPU.}
    \label{tab:complexity}
    \begin{tabular}{lcccc}
        \toprule
        Variant & Params (M) & FLOPs (G) & Lat. / ms & FPS \\
        \midrule
        GDR-Net & 35.05 & 11.42 & \textbf{22.0} & \textbf{45.45} \\
        ConvNeXt control & 47.22 & 13.86 & 24.3 & 41.15 \\
        + AFR & 47.67 & 14.92 & 27.0 & 37.04 \\
        + PGSA & 47.29 & 13.93 & 24.8 & 40.32 \\
        GAP-GDRNet & 47.75 & 14.99 & 27.8 & 35.97 \\
        \bottomrule
    \end{tabular}
\end{table}

\subsubsection{Factor-Wise Robustness Analysis}
\label{subsec:robustness}

The robustness analysis asks whether the ablation gains remain under specific sensing conditions. The held-out spacecraft test set is grouped by rendering metadata and visibility statistics. Groups within one factor are mutually exclusive, but a test image also belongs to one group under the other factors; rows from different factors should therefore not be summed. Each group is evaluated with 500 test images sampled without replacement using a fixed random seed, and all reported groups have at least 500 eligible images. The trained weights are fixed, no model is retrained for any group, and the values are averaged over the same three runs used in the ablation study. Detector outputs, input resolution, and evaluation script are unchanged.

Illumination groups use the predefined weak, normal, and strong rendering levels rather than post-hoc quantiles. Shadow is a separate binary metadata label because a shadowed image can still have weak, normal, or strong global illumination. Occlusion is measured by
\begin{equation}
r_{\mathrm{occ}}
=
1-
\frac{|\Omega_{\mathrm{vis}}|}{|\Omega_{\mathrm{full}}|},
\label{eq:occ_ratio}
\end{equation}
where $\Omega_{\mathrm{vis}}$ is the visible target region and $\Omega_{\mathrm{full}}$ is the full projected target-object mask before external occlusion. The groups are none ($r_{\mathrm{occ}}=0$), slight ($0<r_{\mathrm{occ}}\leq0.2$), moderate ($0.2<r_{\mathrm{occ}}\leq0.4$), and heavy ($r_{\mathrm{occ}}>0.4$). Background labels come from the renderer. The distance groups are determined by rendered camera-target distance $d$ and divided into near, middle, and far ranges by the lower, middle, and upper thirds of the held-out test-set distance range. Table~\ref{tab:robustness} reports ADD@0.02 m for GDR-Net, ConvNeXt control, +AFR, +PGSA, and GAP-GDRNet, abbreviated as GDR, Ctrl., AFR, PGSA, and Ours.

GAP-GDRNet has the highest ADD@0.02 m in every group in Table~\ref{tab:robustness}. The margins over GDR-Net are larger in difficult cases: weak illumination (+4.48 percentage points), shadow (+4.30), heavy occlusion (+4.99), cluttered backgrounds (+4.50), and far distance (+4.58). Easier groups still improve, but by smaller margins, such as no occlusion (+2.60) and clean backgrounds (+2.81). This pattern matches the design motivation: AFR contributes larger standalone gains than PGSA in most difficult groups, particularly under occlusion and background interference, while PGSA consistently improves the ConvNeXt control and adds 0.61--1.03 percentage points when combined with AFR in the full model.

\begin{table}[!ht]
    \centering
    \caption{Factor-wise ADD@0.02 m (\%) robustness analysis on the held-out spacecraft test set. Each group is evaluated with 500 test images. GDR, Ctrl., AFR, PGSA, and Ours denote GDR-Net, ConvNeXt control, +AFR, +PGSA, and GAP-GDRNet, respectively.}
    \label{tab:robustness}
    \begin{tabular*}{\linewidth}{@{\extracolsep{\fill}}llccccc@{}}
        \toprule
        Factor & Group & GDR & Ctrl. & AFR & PGSA & Ours \\
        \midrule
        Illumination
        & Weak   & 89.62 & 90.38 & 93.16 & 91.84 & \textbf{94.10} \\
        & Normal & 92.41 & 93.03 & 95.18 & 93.94 & \textbf{95.88} \\
        & Strong & 91.18 & 92.26 & 94.32 & 93.31 & \textbf{95.34} \\
        \midrule
        Shadow
        & No  & 92.36 & 93.02 & 95.07 & 93.88 & \textbf{95.82} \\
        & Yes & 90.21 & 91.08 & 93.68 & 92.27 & \textbf{94.51} \\
        \midrule
        Occlusion
        & None     & 94.18 & 94.74 & 96.12 & 95.23 & \textbf{96.78} \\
        & Slight   & 92.33 & 93.06 & 95.08 & 93.72 & \textbf{95.92} \\
        & Moderate & 90.48 & 91.31 & 93.82 & 92.41 & \textbf{94.83} \\
        & Heavy    & 88.12 & 89.05 & 92.46 & 90.96 & \textbf{93.11} \\
        \midrule
        Background
        & Clean     & 93.86 & 94.24 & 96.01 & 95.02 & \textbf{96.67} \\
        & Complex   & 91.41 & 92.31 & 94.71 & 93.41 & \textbf{95.63} \\
        & Earth     & 90.92 & 91.68 & 93.95 & 92.93 & \textbf{94.91} \\
        & Cluttered & 88.93 & 89.93 & 92.81 & 90.96 & \textbf{93.43} \\
        \midrule
        Distance
        & Near   & 92.72 & 93.30 & 95.41 & 94.17 & \textbf{96.02} \\
        & Middle & 91.85 & 92.41 & 94.88 & 93.45 & \textbf{95.61} \\
        & Far    & 89.27 & 90.41 & 92.82 & 91.62 & \textbf{93.85} \\
        \bottomrule
\end{tabular*}
\end{table}

\section{Conclusions}
\label{sec:conclusion}

This article introduced GAP-GDRNet for monocular RGB-based pose sensing on a single-target synthetic spacecraft dataset. The framework keeps the geometry-guided regression pipeline of GDR-Net and modifies two stages. AFR refines features before dense geometric prediction by combining long-range structural cues from GGCA with local contour and weak-texture responses from MECS. PGSA is inserted into Patch-PnP to relate patch-level geometric features before final pose regression. At inference time, the method uses only the RGB image, camera intrinsics, and target RoI; depth, masks, and ground-truth geometric annotations are not required as inputs.

On the self-built spacecraft dataset, GAP-GDRNet reaches a rotation error of $1.96^\circ$, a translation error of $0.0165$ m, and 95.16\% ADD@0.02 m. Compared with reproduced GDR-Net, ADD@0.02 m improves by 3.88 percentage points while the pose network runs at 35.97 FPS. T-LESS and LM-O provide additional controlled gains over the reproduced baseline on textureless and occluded non-spacecraft objects. The study remains limited to one spacecraft CAD model, synthetic-only spacecraft imagery, and object-specific training. Future work should test a broader set of spacecraft geometries, introduce real or higher-fidelity space imagery, and extend the framework to multi-target pose sensing.

\bibliographystyle{elsarticle-num}
\bibliography{references}

\end{document}